%% file: AnonymousSubmission2027.tex
\documentclass[letterpaper]{article} 
\usepackage[preprint]{aaai2027}  
\usepackage[hyphens]{url}  
\usepackage{graphicx} 
\usepackage{natbib}  
\usepackage{caption} 
\usepackage{algorithm}
\usepackage{algorithmic}

\usepackage{newfloat}
\usepackage{listings}
\DeclareCaptionStyle{ruled}{labelfont=normalfont,labelsep=colon,strut=off} 
\floatstyle{ruled}
\newfloat{listing}{tb}{lst}{}
\floatname{listing}{Listing}

\usepackage{booktabs}
\usepackage{array}
\usepackage{amsmath}
\usepackage{amssymb}
\newcolumntype{L}[1]{>{\raggedright\arraybackslash}p{#1}}
\newcolumntype{C}[1]{>{\centering\arraybackslash}p{#1}}
\providecommand{\unicode}[1]{}

\usepackage{arydshln}
\usepackage{graphicx}
\usepackage{multirow}
\usepackage{booktabs}
\usepackage[table]{xcolor}
\usepackage{tabularx}
\PassOptionsToPackage{table}{xcolor}
\expandafter\let\csname endtabular*\endcsname\endtabular

\definecolor{gain}{RGB}{0,135,90}
\definecolor{propcolor}{RGB}{31,78,121}
\definecolor{solvercolor}{RGB}{170,35,35}
\definecolor{gain}{RGB}{0,135,90}
\definecolor{cost}{RGB}{170,35,35}
\newcommand{\base}[1]{\makebox[2.8em][l]{#1}}
\newcommand{\score}[2]{\makebox[2.8em][l]{\textbf{#1}}}
\newcommand{\avgscore}[2]{%
  \makebox[4.6em][l]{\textbf{#1} \textcolor{gain}{(+#2)}}%
}
\newcommand{\callscore}[2]{%
  \makebox[4.6em][l]{\textbf{#1} \textcolor{cost}{(#2)}}%
}
\newcommand{\RQ}[2]{%
    \paragraph{RQ#1:}\textbf{\emph{#2}}\enspace
}
\title{Heterogeneous Co-Evolution Learning for Tool-Efficient \\ Search Agents with Zero Labels}
\author {
    Qin-Wen Luo, Sheng Ren, Xiang Chen, Rui Liu, Jun Fang, Naiqiang Tan, Jun Wang, Hua Chai, Sheng-Jun Huang
}
\affiliations {
}
\author{
    Qin-Wen Luo\textsuperscript{\rm 1}, Sheng Ren\textsuperscript{\rm 1}, Xiang Chen\textsuperscript{\rm 1}\corresponding, Rui Liu\textsuperscript{\rm 2}, Jun Fang\textsuperscript{\rm 2}, \\ Naiqiang Tan\textsuperscript{\rm 2}, 
    Jun Wang\textsuperscript{\rm 2},
    Hua Chai\textsuperscript{\rm 2}, Sheng-Jun Huang\textsuperscript{\rm 1}\corresponding 
}
\affiliations{
    \textsuperscript{\rm 1}Nanjing University of Aeronautics and Astronautics, Nanjing, China\\
    \textsuperscript{\rm 2}Didichuxing Co. Ltd, Beijing, China \\
    \texttt{\{luoqw8,xiang\_chen,huangsj\}@nuaa.edu.cn}
}

\begin{document}

\maketitle

\begin{abstract}
Deep search agents extend language models by interleaving reasoning with external retrieval. However, effective search-agent training requires addressing two challenges beyond final-answer optimization. First, the capability frontier of agents is largely determined by training questions, whereas fixed or labeled task collections cannot adapt as agents improve. Second, outcome-only reinforcement learning treats all successful trajectories equally, even when some correct answers rely on redundant searches, creating an inherent tension between answer accuracy and tool efficiency. To address these issues, we introduce \textsc{ZEST-Search}, a zero-label and tool-efficient framework for search-agent training via heterogeneous co-evolution. Starting from unlabeled reference questions, \textsc{ZEST-Search} constructs validated search tasks online through an answer-first and evidence-path-guided generation process. A heterogeneous proposer adaptively constructs training tasks around the solver's evolving capability boundary, while the solver learns search-augmented reasoning through reinforcement learning. To jointly optimize task difficulty and search efficiency, \textsc{ZEST-Search} employs an intermediate-difficulty reward for the proposer and a task-local tool-aware reward for the solver, encouraging informative tasks and correct answers with fewer redundant searches. Experiments on seven general-domain and multi-hop question answering benchmarks demonstrate that \textsc{ZEST-Search} consistently improves search-agent performance while achieving higher tool efficiency without labeled search-training data or expert search trajectories. Further analyses show that heterogeneous generation, adaptive difficulty control, and tool-aware optimization enable more effective search-agent training.
\end{abstract}

\input{section/introduction}
\input{section/preliminaries}
\input{section/methodology}
\input{section/experiments}
\input{section/related_work}
\input{section/conclusion}

\bibliography{aaai2027}

\clearpage
\input{section/appendix}

\end{document}

%% file: section/introduction.tex

\section{Introduction}

Deep search agents tackle challenging questions by interleaving reasoning with external retrieval. Unlike standard retrieval-augmented generation, they must determine not only which evidence to use, but also when and how to search.
Retrieval-augmented models established external evidence as central to knowledge-intensive question answering~\cite{guu2020realm,karpukhin2020dpr,lewis2020rag}, while multi-hop benchmarks highlighted the need to integrate evidence across documents and entities~\cite{yang2018hotpotqa,trivedi2022musique}.
More recent tool-using and retrieval-aware agents integrate search into the reasoning loop, enabling adaptive tool use during inference~\cite{nakano2021webgpt,yao2023react,asai2024selfrag,trivedi2023ircot}.
Accordingly, training search agents requires optimizing both final-answer correctness and search behavior to produce evidence-grounded trajectories.

Reinforcement learning (RL) has become a prominent approach for training such search behavior. 
Early studies train language models to invoke search during reasoning using outcome-based rewards derived from final-answer correctness~\cite{jin2025searchr1,song2025r1searcher,chen2025research}.
Subsequent studies further explore search-cost modeling, stepwise feedback, and data-efficient RL for search agents~\cite{sun2025zerosearch,zheng2025stepsearch,jiang2025s3}.
Despite this progress, existing methods still fail to jointly control two key determinants of search-agent learning. First, training questions largely define the frontier of learnable capabilities. A fixed question set can become a static curriculum, preventing RL from exposing new agent weaknesses. Second, outcome-only rewards cannot distinguish concise successful search trajectories from redundant ones. Consequently, optimizing answer correctness alone may improve accuracy while increasing tool-use costs.

To address the first limitation, recent self-play methods formulate learning as an adaptive proposer--solver game~\cite{zhao2025absolutezero,chen2025sqlm, serl, beyond}.
In search-agent training, the proposer generates verifiable questions, while the solver answers them through search.
The performance of the solver then guides proposer updates, allowing task difficulty to evolve with the solver throughout training~\cite{lu2025search,yue2026drzero,wu2026knowledgegraph}.
However, task generation is typically constrained by fixed predefined sources, such as answer pools, limiting exploration beyond the initial task space.
Assigning both roles to the same model further couples online generation cost to solver scale and limits signal complementarity, since shared knowledge priors can confine generated questions to familiar regions.
This coupling is especially problematic early in training, when base models may produce fewer valid and informative questions, yielding fewer effective training groups and slower optimization.
Moreover, adaptive task generation alone does not resolve the second limitation and may even exacerbate tool inefficiency. As generated tasks become harder, a solver optimized solely for final-answer correctness may resort to increasingly extensive search, even when additional queries provide little marginal value.

To address these limitations, we propose \textsc{ZEST-Search}, a
heterogeneous proposer--solver co-evolution framework for
\textbf{Z}ero-label, tool-\textbf{E}fficient
\textbf{S}earch-agent \textbf{T}raining.
Rather than sampling answers from a fixed pool, the proposer first
generates target entities conditioned on unlabeled reference questions.
Each target is validated for semantic compatibility, entity type,
extractability, and retrieval support.
\textsc{ZEST-Search} then constructs source-supported multi-hop evidence
paths terminating at the validated targets and conditions question
generation on these paths. Programmatic verification filters malformed,
answer-leaking, and insufficiently supported tasks before solver
training. These target-validation, evidence-construction, and
task-filtering stages rely only on lightweight retrieval and NLP
components, requiring no additional LLM inference.
Decoupling task generation from solving enables lightweight or
cross-family proposers to generate complementary tasks around the
solver's evolving capability boundary. 
Role-specific objectives coordinate their co-evolution: an intermediate-difficulty reward steers the proposer toward validated questions near the solver's learning frontier, while a task-local tool-aware reward favors correct trajectories with lower relative search costs. 
Together, these components expand the training space beyond predefined answer pools and enable zero-label learning of evidence-grounded and tool-efficient search behavior.

Our contributions are summarized as follows:
\begin{itemize}
    \item We propose a zero-label, answer-first pipeline that constructs
    validated target entities, source-supported multi-hop evidence paths,
    and verifiable search tasks without labeled QA pairs or expert search
    trajectories.

    \item By formulating search-agent training as heterogeneous proposer--solver co-evolution, our framework enables lightweight or
    cross-family proposers to generate complementary tasks around the
    solver's evolving frontier.

    \item Role-specific objectives jointly optimize task difficulty and
    search efficiency, favoring informative questions near the solver's
    learning frontier and correct trajectories with fewer redundant
    searches.
\end{itemize}

%% file: section/preliminaries.tex

\section{Preliminaries}
\label{preliminaries}

\paragraph{Reinforcement Learning with Verifiable Rewards.}
Reinforcement learning with verifiable rewards (RLVR) trains language models using outcomes that can be checked by programmatic verifiers~\cite{shao2024deepseekmath, yu2025dapo}.
Let $x$ denote an input prompt and $a^\star$ denote its target answer. 
A policy $\pi_\theta$ generates the response $y=(o, a)$, where $o$ is the reasoning trace, and $a$ is the final answer. 
A verifier assigns the reward $r(x,y)=\mathbb{I}(a=a^\star)$.

Group Relative Policy Optimization (GRPO)~\cite{shao2024deepseekmath} leverages group-wise advantage estimation to enable critic-free policy optimization.
For a prompt-target pair $(x,a^\star)$, the policy generates $K$ responses $\mathcal{Y}=\{y^1,\ldots,y^K\}$, where $y^k=(o^k,a^k)$, and obtains rewards by matching final answers with the target answers.
The group-relative advantages are computed by normalizing rewards with group statistics:
\begin{equation}
    A^k =
    \frac{r^k-\mu_R}{\sigma_R+\xi},
\end{equation}
where $\mu_R$ and $\sigma_R$ denote the mean and standard deviation of rewards in $\mathcal{Y}$, and $\xi$ avoids division by zero.
Denote that $\rho^k(\theta)=\pi_\theta(y^k\mid x)/\pi_{\theta_{\mathrm{old}}}(y^k\mid x)$ is the importance sampling ratio. GRPO optimizes the clipped objective
\begin{equation}
    J_{\mathrm{GRPO}}(\theta)
    =
    \mathbb{E}_{x,\mathcal{Y}}
    \left[
    A_\mathrm{IS}^k
    -\alpha \mathbb{D}_{\mathrm{KL}}
    \big(\pi_\theta \,\|\, \pi_{\mathrm{ref}}\big)
    \right],
\end{equation}
where
\begin{equation}
\begin{aligned}
    A_\mathrm{IS}^k
    =
    \min\big[&
    \rho^k(\theta)A^k,\,
    \mathrm{clip}(\rho^k(\theta),1-\epsilon,1+\epsilon)
    A^k
    \big],
\end{aligned}
\end{equation}
$\epsilon$ controls the trust region and $\alpha$ weights the reference-policy regularizer.

\paragraph{Search-Agent Rollouts}
A search agent interacts with a retrieval environment during response generation. Given the input prompt $x$, the agent maintains a history
\begin{equation}
    h_t=(x,u_{<t},z_{<t}),
\end{equation}
where $u_{<t}$ are generated segments and $z_{<t}$ are retrieved observations. At step $t$, the policy emits either a normal reasoning segment, a search action, or a final-answer action:
\begin{equation}
    u_t \sim \pi_\theta(\cdot \mid h_t).
\end{equation}
If $u_t$ is a valid parsed search action with query $s_t$, the environment returns
\begin{equation}
    z_t = \mathcal{R}(s_t;k_{\mathrm{ret}}),
\end{equation}
where $\mathcal{R}$ is the retriever and $k_{\mathrm{ret}}$ is the top-$k$ retrieval depth. If $u_t$ is a final-answer action, the response terminates with final answer $a$. A complete response keeps the same notation as above:
\begin{equation}
    y=(o,a),
    \qquad
    o=(u_1,z_1,\ldots,u_T,z_T).
\end{equation}
Only valid parsed search actions are counted as tool calls:
\begin{equation}
    N(y)=
    \sum_{t=1}^{T}
    \mathbb{I}_{\mathrm{search}}(u_t).
\end{equation}
Invalid tool syntax is kept in the textual trajectory but does not increase $N(y)$. Retrieved observations condition later actions, but they are not policy-generated tokens; therefore, they are masked in the actor objective.

%% file: section/methodology.tex

\begin{figure*}[t]
\centering
\includegraphics[width=\textwidth]{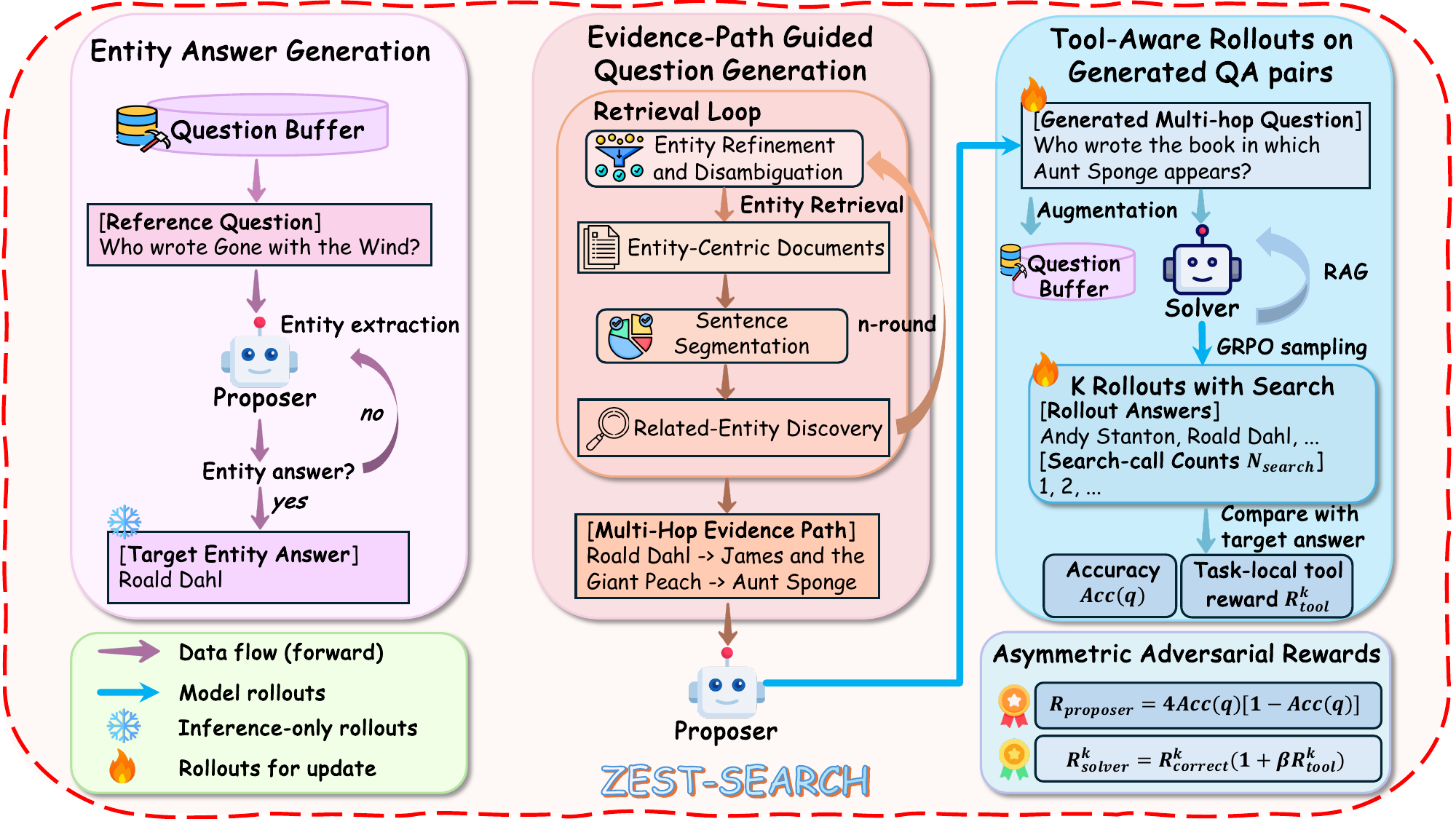}
\caption{Overview of \textsc{ZEST-Search}. From left to right, an unlabeled reference question provides a semantic template for target-entity generation; retrieved evidence is organized into a source-supported path for generating a new question; and a heterogeneous solver produces grouped search rollouts that yield role-specific rewards for proposer--solver co-evolution.}
\label{fig:zest-method}
\end{figure*}

\section{Methodology}
\label{sec:method}

\subsection{Framework Overview}
\textsc{ZEST-Search} trains search agents through an answer-anchored task generation and optimization process, as shown in Figure~\ref{fig:zest-method}.
Given an unlabeled reference question, the proposer generates a target entity for a new search task.
From this target, the framework retrieves entity-centric documents, identifies sentences mentioning queried entities, and extracts co-occurring entities.
Recursively querying selected entities yields a source-traceable multi-hop evidence path.
Conditioned on the target and evidence context, the proposer generates a multi-hop question whose answer is the target.
After validation, the accepted QA pairs are passed to the solver for grouped search rollouts and GRPO optimization.
A verifier compares each final answer against the target, while valid search calls quantify tool use.
These signals define role-specific rewards: the proposer is encouraged
to generate tasks of intermediate difficulty, while the solver
is rewarded for correct trajectories with lower relative search cost.

\subsection{Entity Answer Generation}
\label{sec:method-answer}
To avoid reliance on a fixed data source, we derive target
entities online from abundant unlabeled questions. 
Given a reference question $x$, the proposer determines whether its expected answer is entity-valued. 
If so, it generates an entity of the implied semantic type. 
Otherwise, it identifies a salient entity in $x$ and generates an entity of the same type.
This procedure maps both cases to entity-valued targets suitable for entity-centric retrieval, without reference answers.

Formally, the proposer samples
\begin{equation}
    u^{a}
    \sim
    \pi_{p}^{\phi}
    \bigl(
        \cdot
        \mid
        P_{a}(x)
    \bigr),
    \qquad
    \widetilde{a}
    =
    E_{a}(u^{a}),
\end{equation}
where $P_a$ is the target-entity generation prompt and $E_a$ extracts the
candidate entity from the structured output. The prompt requests one
concrete, distinct, and retrievable entity rather than a direct answer
to $x$, avoiding the complex and error-prone process of answer inference.
The candidate $\widetilde{a}$ is then screened by a lightweight
validation pipeline based on deterministic parsing, entity recognition, and answer-side retrieval verification. These non-generative checks require no additional LLM inference and efficiently filter out malformed outputs and non-entity candidates. 
Candidates passing all checks are accepted as $a^\star$. 
The validated target anchors evidence-path construction and serves as the verification target for the generated search task.

\subsection{Evidence-Path Guided Question Generation}
\label{sec:method-question}

Given a retrieval-supported target entity $a^{\star}$, we construct a
source-grounded evidence context for multi-hop question generation. 
We treat $a^{\star}$ as the answer node of a compact evidence graph and initialize the document pool with retrieval results obtained for $a^{\star}$.

At each hop, each selected frontier entity is normalized and disambiguated before retrieval. 
Retrieved documents are segmented into sentences, and those mentioning both the selected entity and another valid entity are retained as evidence. 
Each co-occurring entity defines an evidence-backed edge and becomes a frontier candidate for the next hop. 
This process recursively expands the frontier until no valid candidate remains or the maximum hop budget is reached.
An evidence path from an outer clue to $a^{\star}$ is then constructed:
\begin{equation}
    p =
    \left(
        v_0
        \rightarrow
        v_1
        \rightarrow
        \cdots
        \rightarrow
        a^{\star}
    \right),
\end{equation}
where the entities are ordered toward the target answer.
Each edge is paired with its source document and supporting sentence.
Paths with repeated entities, discontinuous transitions, or insufficient evidence are discarded. 
Remaining candidates are ranked by evidence quality and retrieval relevance, and the top-ranked path is selected.
The resulting evidence context is represented as
\begin{equation}
    e =
    \left(
        p,\mathcal{S},\mathcal{D}
    \right),
\end{equation}
where $\mathcal{S}$ contains the supporting sentences and $\mathcal{D}$ contains the corresponding documents.
The proposer then samples a question conditioned on the target and
evidence context:
\begin{equation}
    u^{q}
    \sim
    \pi_{p}^{\phi}
    \left(
        \cdot
        \mid
        P_q(a^{\star},e)
    \right),
    \qquad
    q = E_q(u^{q}),
\end{equation}
where $P_q$ is the evidence-conditioned generation prompt and $E_q$
extracts the question from the structured output. 
The prompt instructs the proposer to conceal the target while preserving the relational clues encoded by the path, yielding a multi-hop search task with the intended answer $a^{\star}$.

Evidence path construction uses only lightweight retrieval and NLP components, while the proposer is invoked for question generation.
Validated questions are then added to the reference pool to support later rounds and progressively expand the training distribution.

\subsection{Tool-Aware Rollouts on Generated QA Pairs}
\label{sec:method-rollout}
Given a validated task $(q,a^{\star})$, the solver interacts with the
retrieval environment according to the rollout protocol. 
For each task, we sample a group of $K$ independent
trajectories:
\begin{equation}
    \{y^{k}\}_{k=1}^{K}
    \sim
    \pi_{s}^{\theta}(\cdot\mid q).
\end{equation}
Each rollout terminates when the solver produces a valid final answer
or exhausts the search-turn budget.
The final answer $a^k$ of rollout $k$ is compared with the target
$a^\star$ to obtain a verifiable exact-match reward:
\begin{equation}
    R_{\mathrm{correct}}^k
    =
    \mathbb{I}_{\mathrm{EM}}
    \left[
        a^{k}=a^{\star}
    \right].
\end{equation}
The grouped rollouts yield the task-level solver accuracy
\begin{equation}
    \mathrm{Acc}(q)
    =
    \frac{1}{K}
    \sum_{k=1}^{K}
    R_{\mathrm{correct}}^k,
\end{equation}
together with the per-rollout search-call counts
$\{N(y^k)\}_{k=1}^{K}$. The former characterizes the task's current
difficulty for the solver, whereas the latter quantifies
trajectory-level tool use. These task- and trajectory-level statistics
are subsequently used to define the role-specific rewards for proposer
and solver optimization.

\subsection{Asymmetric Adversarial Rewards}
\label{sec:method-reward}
\textsc{ZEST-Search} formulates search-agent training as heterogeneous proposer–solver co-evolution. The proposer generates tasks around the solver’s evolving capability boundary, while the solver learns from these tasks. 
Unlike self-play methods, the two roles can be instantiated with heterogeneous models.
We therefore use an intermediate-difficulty reward for the proposer to encourage informative tasks and a task-local tool-aware reward for the solver to favor correct answers with less unnecessary tool use.

\paragraph{Intermediate-Difficulty Reward for the Proposer.}
The proposer should generate questions that remain informative for the current solver. 
Prior curriculum-learning studies suggest that questions with intermediate pass rates provide stronger optimization signals than those almost always solved or missed~\cite{mmr1,vcrl,crues,rprc}. 
We therefore define the proposer reward using the grouped solver accuracy
$\mathrm{Acc}(q)$:
\begin{equation}
\label{eq:proposer-reward}
    R_{\mathrm{proposer}}
    =
    4\mathrm{Acc}(q)
    \left[
        1-\mathrm{Acc}(q)
    \right].
\end{equation}
Invalid questions receive a fixed negative reward
$\beta_{\mathrm{invalid}}<0$.
For valid questions, Eq.~\ref{eq:proposer-reward} is maximized at
$\mathrm{Acc}(q)=0.5$ and vanishes for all-correct or all-wrong groups,
thereby favoring tasks near the solver's current learning frontier.

A monotonic alternative, $1-\mathrm{Acc}(q)$, assigns its maximum reward
when all solver rollouts fail. However, complete failure may reflect an
incorrect target, an ambiguous question, or insufficient evidence rather
than meaningful difficulty.
By assigning zero reward to all-wrong groups,
the intermediate-difficulty reward avoids this failure mode.

\paragraph{Tool-Aware Reward for the Solver.}
External search should be used only when the solver's parametric
knowledge and available evidence are insufficient. Unnecessary
retrieval increases inference cost and may introduce irrelevant
information.
For each task, we define a task-local tool reward by normalizing search
calls across its $K$ rollouts:
\begin{equation}
\label{eq:tool-reward}
    R_{\mathrm{tool}}^k
    =
    -
    \frac{
        N(y^k)
    }{
        \max_{\ell \leq K} N(y^\ell)+1
    }.
\end{equation}
The reward is non-positive, so trajectories with fewer search calls
receive smaller penalties. Because normalization is performed within each
task, it accounts for question-specific retrieval demands and is
naturally compatible with GRPO.
The final solver reward is
\begin{equation}
\label{eq:solver-reward}
    R_{\mathrm{solver}}^{k}
    =
    R_{\mathrm{correct}}^k
    \left(
        1+
        \lambda R_{\mathrm{tool}}^k
    \right),
\end{equation}
where $\lambda>0$ controls the tool-cost discount. 
Multiplying the tool reward by correctness prevents incorrect rollouts from benefiting from less search. Among correct rollouts, fewer search calls receive higher rewards, encouraging the solver to preserve accuracy while selectively using retrieval.

%% file: section/experiments.tex
\begin{table*}[t]
\centering

{\small
\setlength{\tabcolsep}{1.5pt}

\begin{tabular}{l@{\hspace{15pt}}l@{\hspace{10pt}}*{9}{l}}
\toprule

\multicolumn{2}{l}{}
& \multicolumn{3}{c}{\textbf{GeneralQA}}
& \multicolumn{4}{c}{\textbf{Multi-HopQA}}
&  \\
\cmidrule(lr){3-5}
\cmidrule(lr){6-9}

\multicolumn{1}{c}{\textbf{\textcolor{propcolor}{Proposer} / }\textbf{\textcolor{solvercolor}{Solver}}}
& \multicolumn{1}{c}{\textbf{Method}}
& \multicolumn{1}{c}{NQ}
& \multicolumn{1}{c}{TriviaQA}
& \multicolumn{1}{c}{PopQA}
& \multicolumn{1}{c}{HotpotQA}
& \multicolumn{1}{c}{2Wiki}
& \multicolumn{1}{c}{MuSiQue}
& \multicolumn{1}{c}{Bamboogle}
& \multicolumn{1}{c}{\textbf{Avg$(\uparrow)$}}
& \multicolumn{1}{c}{\textbf{Queries$(\downarrow)$}} \\

\midrule
\rowcolor{gray!25}
\multicolumn{11}{c}{\textbf{Zero-Label Training from Base Models}} \\
\midrule

\textcolor{propcolor}{Qwen2.5-3B-Instruct /}
& Solver Base
& \base{16.7}
& \base{30.7}
& \base{17.8}
& \base{12.4}
& \base{7.2}
& \base{2.3}
& \base{6.4}
& \base{13.4}
& \base{0.85} 
\\

\textcolor{solvercolor}{Qwen2.5-3B-Base}
& \textbf{Ours}
& \score{36.0}{}
& \score{57.0}{}
& \score{38.2}{}
& \score{26.0}{}
& \score{25.4}{}
& \score{5.3}{}
& \score{7.2}{}
& \avgscore{27.9}{14.5}
& \callscore{\textcolor{black!65}{\underline{1.01}}}{\underline{+0.16}} \\

\noalign{\vskip 1pt}
\cdashline{1-11}
\noalign{\vskip 1pt}

\textcolor{propcolor}{Qwen2.5-7B-Base /}
& Solver Base
& \base{25.8}
& \base{43.3}
& \base{28.3}
& \base{15.8}
& \base{9.2}
& \base{6.4}
& \base{24.1}
& \base{21.8}
& \base{1.59} \\

\textcolor{solvercolor}{Qwen2.5-7B-Base}
& \textbf{Ours}
& \score{37.1}{}
& \score{61.9}{}
& \score{40.3}{}
& \score{34.7}{}
& \score{31.8}{}
& \score{11.7}{}
& \score{32.0}{}
& \avgscore{35.6}{13.8}
& \callscore{1.17}{-0.42} \\

\noalign{\vskip 1pt}
\cdashline{1-11}
\noalign{\vskip 1pt}

\textcolor{propcolor}{Qwen2.5-7B-Instruct /}
& Solver Base
& \base{25.8}
& \base{43.3}
& \base{28.3}
& \base{15.8}
& \base{9.2}
& \base{6.4}
& \base{24.1}
& \base{21.8}
& \base{1.59} \\

\textcolor{solvercolor}{Qwen2.5-7B-Base}
& \textbf{Ours}
& \score{40.2}{}
& \score{64.5}{}
& \score{43.8}{}
& \score{36.3}{}
& \score{35.4}{}
& \score{12.0}{}
& \score{35.2}{}
& \avgscore{38.2}{16.4}
& \callscore{1.16}{-0.43} \\

\noalign{\vskip 1pt}
\cdashline{1-11}
\noalign{\vskip 1pt}







\midrule
\rowcolor{gray!25}
\multicolumn{11}{c}{\textbf{Scaling to Larger Solvers with Lightweight Proposers}} \\
\midrule

\textcolor{propcolor}{Qwen2.5-3B-Base /}
& Solver Base
& \base{25.8}
& \base{43.3}
& \base{28.3}
& \base{15.8}
& \base{9.2}
& \base{6.4}
& \base{24.1}
& \base{21.8}
& \base{1.59} \\

\textcolor{solvercolor}{Qwen2.5-7B-Base}
& \textbf{Ours}
& \score{39.1}{}
& \score{61.9}{}
& \score{41.9}{}
& \score{33.7}{}
& \score{31.7}{}
& \score{9.9}{}
& \score{29.6}{}
& \avgscore{35.4}{13.6}
& \callscore{1.29}{-0.30} 
\\

\noalign{\vskip 1pt}
\cdashline{1-11}
\noalign{\vskip 1pt}

\textcolor{propcolor}{Qwen2.5-3B-Instruct /}
& Solver Base
& \base{25.8}
& \base{43.3}
& \base{28.3}
& \base{15.8}
& \base{9.2}
& \base{6.4}
& \base{24.1}
& \base{21.8}
& \base{1.59} \\

\textcolor{solvercolor}{Qwen2.5-7B-Base}
& \textbf{Ours}
& \score{38.6}{}
& \score{61.5}{}
& \score{41.6}{}
& \score{32.2}{}
& \score{30.3}{}
& \score{11.7}{}
& \score{36.0}{}
& \avgscore{36.0}{14.2}
& \callscore{1.17}{-0.42} 
\\

\midrule
\rowcolor{gray!25}
\multicolumn{11}{c}{\textbf{Generalization Across Model Families}} \\
\midrule

\textcolor{propcolor}{Qwen3-4B /}
& Solver Base
& \base{25.8}
& \base{43.3}
& \base{28.3}
& \base{15.8}
& \base{9.2}
& \base{6.4}
& \base{24.1}
& \base{21.8}
& \base{1.59} \\

\textcolor{solvercolor}{Qwen2.5-7B-Base}
& \textbf{Ours}
& \score{40.0}{}
& \score{62.9}{}
& \score{43.0}{}
& \score{34.3}{}
& \score{33.5}{}
& \score{11.8}{}
& \score{32.8}{}
& \avgscore{36.9}{15.1}
& \callscore{1.08}{-0.51} \\

\noalign{\vskip 1pt}
\cdashline{1-11}
\noalign{\vskip 1pt}

\textcolor{propcolor}{LLaMA3.2-3B-Instruct /}
& Solver Base
& \base{26.7}
& \base{46.4}
& \base{29.4}
& \base{20.6}
& \base{16.6}
& \base{9.5}
& \base{31.2}
& \base{25.8}
& \base{1.60} \\

\textcolor{solvercolor}{LLaMA3-8B-Instruct}
& \textbf{Ours}
& \score{38.6}{}
& \score{64.3}{}
& \score{40.6}{}
& \score{33.0}{}
& \score{28.4}{}
& \score{12.1}{}
& \score{41.6}{}
& \avgscore{36.9}{11.1}
& \callscore{1.50}{-0.10} \\

\midrule
\rowcolor{gray!25}
\multicolumn{11}{c}{\textbf{Adaptation of Search-Specialized Agents}} \\
\midrule

\textcolor{propcolor}{Qwen2.5-7B-Instruct /}
& Solver Base
& \base{30.1}
& \base{45.4}
& \base{29.8}
& \base{23.1}
& \base{12.9}
& \base{8.2}
& \base{27.2}
& \base{25.2}
& \base{2.26} 
\\

\textcolor{solvercolor}{Search-R1}
& \textbf{Ours}
& \score{34.4}{}
& \score{50.4}{}
& \score{34.0}{}
& \score{28.3}{}
& \score{22.6}{}
& \score{10.6}{}
& \score{32.0}{}
& \avgscore{30.3}{4.9}
& \callscore{1.58}{-0.68} \\

\noalign{\vskip 1pt}
\cdashline{1-11}
\noalign{\vskip 1pt}

\textcolor{propcolor}{Qwen2.5-7B-Instruct /}
& Solver Base
& \base{35.1}
& \base{56.9}
& \base{35.9}
& \base{29.6}
& \base{30.6}
& \base{9.5}
& \base{25.0}
& \base{31.8}
& \base{1.12}
\\

\textcolor{solvercolor}{ZeroSearch}
& \textbf{Ours}
& \score{41.7}{}
& \score{65.1}{}
& \score{41.9}{}
& \score{33.1}{}
& \score{33.2}{}
& \score{10.3}{}
& \score{\textcolor{black!65}{\underline{24.8}}}{}
& \avgscore{35.7}{3.9}
& \callscore{1.09}{-0.03} \\

\bottomrule
\end{tabular}
}

\caption{Main experimental results. Results above the base model are in bold, while those below it are underlined.}
\label{tab:main_results}
\end{table*}

\section{Experiments}
\label{sec:experiments}

\begin{figure*}[t]
    \centering
    \includegraphics[width=0.32\textwidth]{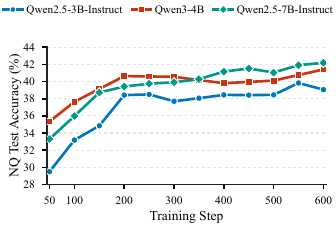}
    \hfill
    \includegraphics[width=0.32\textwidth]{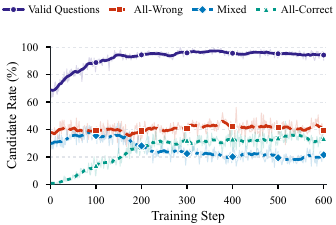}
    \hfill
    \includegraphics[width=0.32\textwidth]{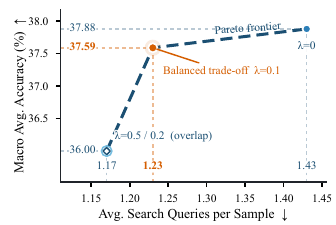}
    \caption{
        Training and efficiency analyses.
        (a) The training dynamics of proposers with different scales.
        (b) Question-generation quality during the training process.
        (c) Accuracy--efficiency trade-off under different tool-cost coefficients.
    }
    \label{fig:analysis_panels}
\end{figure*}

To comprehensively evaluate \textsc{ZEST-Search}, we investigate its effectiveness, scalability across heterogeneous proposer--solver settings, and component contributions.

\subsection{Experimental Setup}
\label{sec:exp_setup}

\paragraph{Benchmarks and Evaluation Metrics.}
We evaluate on three general-domain QA benchmarks: NQ~\cite{kwiatkowski2019natural}, TriviaQA~\cite{joshi2017triviaqa}, and PopQA~\cite{mallen2023popqa}, and four multi-hop benchmarks: HotpotQA~\cite{yang2018hotpotqa}, 2Wiki~\cite{ho2020twowiki},
MuSiQue~\cite{trivedi2022musique}, Bamboogle~\cite{press2023bamboogle}, requiring iterative retrieval and evidence integration.
We report normalized exact match (EM) on each benchmark, the unweighted macro-average EM across all benchmarks, and Queries, defined as the average number of successfully parsed search actions per example.

\paragraph{Training and Evaluation Settings.}
Training uses only question text from an unlabeled reference buffer initialized with questions from the NQ training split. 
Target answers, evidence paths, and QA pairs are constructed online.
Each accepted task receives five solver rollouts, each allowing up to three search turns, with three retrieved passages per query.
The proposer is optimized with REINFORCE++~\cite{hu2025reinforceplusplus} using task-level rewards derived
from grouped solver outcomes, whereas the solver is optimized with GRPO
over tool-interleaved trajectories. 
At evaluation time, we use temperature-zero decoding with three search turns.

\subsection{Main Results}
\label{sec:main_results}

\RQ{1}{Can \textsc{ZEST-Search} improve search-augmented reasoning
without labeled search-training data?}
We compare each solver before and after \textsc{ZEST-Search} training. Since existing self-play search-agent methods often adopt different model APIs, datasets, LLM-as-Judges, or retrieval environments, direct comparisons under identical training budgets remain challenging.
We therefore additionally evaluate \textsc{ZEST-Search} through continual training on search-specialized agents.
Because training uses only unlabeled questions and constructs tasks online, these comparisons assess whether search-augmented reasoning improves without labeled QA pairs or expert trajectories.
As shown in Table~\ref{tab:main_results}, \textsc{ZEST-Search}
substantially improves search-augmented reasoning using only unlabeled questions for task construction.
Across all seven benchmarks, solvers of different scales and model
families consistently benefit from training under diverse proposer
configurations, demonstrating robust gains on both general-domain and
multi-hop question answering. 
The same procedure also improves search-specialized solvers, including Search-R1~\cite{jin2025searchr1} and ZeroSearch~\cite{sun2025zerosearch}, showing that \textsc{ZEST-Search} can further enhance existing search agents.
These results show that \textsc{ZEST-Search} can both develop search-augmented reasoning from base models and further improve agents that have already undergone search-oriented training.
Overall, ZEST-Search improves the performance of search agents without labeled training data.

\RQ{2}{Can \textsc{ZEST-Search} effectively train solvers with lightweight proposers?}
We evaluate whether proposers smaller than the solver can generate effective training tasks, as reported in
Table~\ref{tab:main_results}. Smaller proposers provide
effective training signals for larger solvers. In particular, Qwen3-4B
performs comparably to the larger Qwen2.5-7B proposer, while
Qwen2.5-3B-Instruct also produces strong gains. 
The smaller but newer proposers, such as Qwen3 and LLaMA-3.2 models, can generate higher-quality training tasks, providing more effective supervision for solver improvement.
These results show that online task generation can be delegated to lightweight models without matching the solver's parameter scale. 
Although task quality varies across proposer checkpoints, solvers consistently benefit from both smaller and cross-family proposers.

\RQ{3}{Does \textsc{ZEST-Search} improve the trade-off between answer
accuracy and search cost?}
As shown in Table~\ref{tab:main_results}, \textsc{ZEST-Search} improves
answer accuracy while reducing average search calls for most solvers. 
This indicates that the correctness-gated tool-aware objective encourages selective retrieval while preserving answer accuracy.
Qwen2.5-3B-Base is the only exception, showing a slight increase
in search calls alongside a substantial accuracy gain. Because this
weak base model initially underuses search, the increase indicates more
appropriate tool invocation rather than degraded efficiency. 
After training, the solver learns to better balance parametric knowledge and retrieval.
Overall, \textsc{ZEST-Search} improves the accuracy--cost trade-off by promoting selective search and avoiding redundant queries.

\RQ{4}{How does proposer heterogeneity affect task generation and training dynamics?}
To analyze how proposer heterogeneity affects training, we compare proposers with different model scales, families, and instruction alignment.
First, cross-family proposers such as Qwen3-4B and LLaMA3.2-3B-Instruct introduce complementary knowledge priors and reasoning patterns, enabling larger solvers to benefit from diverse task distributions without requiring the matching scale of the proposer. This indicates that task generation quality is not strictly bounded by solver capacity.
Beyond task diversity, proposer heterogeneity also affects how quickly useful supervision emerges during training. As shown in Figure~\ref{fig:analysis_panels}(a), newer or instruction-tuned proposers generate a higher proportion of valid and informative questions at the beginning of training, providing stronger optimization signals and accelerating solver improvement.
Overall, heterogeneous proposer--solver training decouples task generation cost, knowledge priors, and generation quality from solver capacity.

\begin{table}[t]
    \centering
    \setlength{\tabcolsep}{3.5pt}

    \begin{tabular*}{\columnwidth}{
        @{\extracolsep{\fill}}lcc@{}
    }
        \toprule
        \textbf{Reward Type}
        & $\mathbf{4\mathrm{Acc}[1-\mathrm{Acc}]}$
        & $\mathbf{1-\mathrm{Acc}}$ \\
        \midrule

        \multicolumn{3}{l}{
            \textit{Valid QA Distribution (\%)}
        } \\

        Mixed Questions $\uparrow$
        & \textbf{21.63}
        & 12.81 \\

        All-Wrong Questions $\downarrow$
        & \textbf{43.80}
        & 83.95 \\

        All-Right Questions
        & 34.57
        & 3.24 \\

        \midrule

        \multicolumn{3}{l}{
            \textit{Performance}
        } \\

        Avg. Accuracy $\uparrow$
        & \textbf{37.59}
        & 36.00 \\

        Avg. Queries $\downarrow$
        & 1.23
        & \textbf{1.17} \\

        \bottomrule
    \end{tabular*}

    \caption{
        Effect of proposer rewards on generated task difficulty and solver performance.
    }
    \label{tab:proposer_reward_ablation}
\end{table}

\subsection{Ablation and Analysis}
\label{sec:analysis}

\paragraph{Intermediate-Difficulty Reward Analysis.}
Table~\ref{tab:proposer_reward_ablation} compares the proposed
intermediate-difficulty reward with the monotonic adversarial objective.
The latter over-rewards all-wrong groups, concentrating generation on unsolved questions. In contrast, the intermediate-difficulty reward increases the proportion of mixed groups while reducing all-wrong cases. Mixed groups provide informative relative signals for policy optimization because they contain both successful and failed solver trajectories. Figure~\ref{fig:analysis_panels}(b) further shows that the mixed-group proportion remains stable throughout training. These results indicate that the proposed reward more effectively targets the solver's current learnable boundary.

\begin{table}[t]
    \centering
    \setlength{\tabcolsep}{2.5pt}

    \begin{tabular*}{\columnwidth}{
        @{\extracolsep{\fill}}lccc@{}
    }
        \toprule
        \textbf{Metric}
        & \textbf{Answer Retrieval}
        & \textbf{EPG}
        & \textbf{EPG + RL} \\
        \midrule

        \multicolumn{4}{l}{
            \textit{Question Quality (\%)}
        } \\

        Answer leakage $\downarrow$
        & 9.6
        & 21.2
        & \textbf{2.0} \\

        Valid question $\uparrow$
        & 29.2
        & 27.6
        & \textbf{42.4} \\

        \midrule

        \multicolumn{4}{l}{
            \textit{Hop Distribution among Valid Questions (\%)}
        } \\

        1-hop
        & 95.2
        & 71.7
        & 54.7 \\

        2-hop
        & 4.8
        & 26.1
        & \textbf{37.7} \\

        3-hop
        & 0.00
        & 2.2
        & \textbf{7.6} \\

        \bottomrule
    \end{tabular*}

    \caption{
        Comparison of question quality under answer-retrieval and evidence-path guided (EPG) generation.
    }
    \label{tab:qa_generation_quality}
\end{table}

\paragraph{Tool-Efficient Search Optimization Analysis.}
Figure~\ref{fig:analysis_panels}(c) shows the accuracy--efficiency
trade-off across different tool-cost coefficients with 400 training steps. Without the
tool-aware reward, the solver achieves the highest accuracy but uses the
most searches. A moderate coefficient markedly reduces search calls
with only a minor accuracy loss, providing the best overall balance.
This demonstrates that the objective is not to minimize retrieval, but to optimize the accuracy-efficiency trade-off.
Larger coefficients further suppress retrieval but increasingly degrade
accuracy. These results show that the tool-aware reward provides
controllable search regularization: moderate penalties discourage
redundant retrieval, whereas excessive penalties can suppress searches
necessary for correct answers.

\paragraph{Evidence-Path Guided Generation Analysis.}
Table~\ref{tab:qa_generation_quality} compares direct
answer-retrieval generation (AR), which conditions question generation
on documents containing the target answer, with the proposed
evidence-path-guided generation (EPG). Even before proposer
optimization, EPG yields substantially more multi-hop questions among examples, indicating that explicit evidence paths promote
compositional generation beyond direct answer--document associations.
However, its lower initial validity and higher answer leakage indicate that structural guidance alone is insufficient for generating well-formed tasks. 
After training, EPG improves validity and leakage control while further increasing the proportion of multi-hop questions. Thus, evidence paths provide compositional structure, while proposer optimization converts them into valid and non-leaking QA pairs.

\begin{figure}
    \centering
    \includegraphics[width=\linewidth]{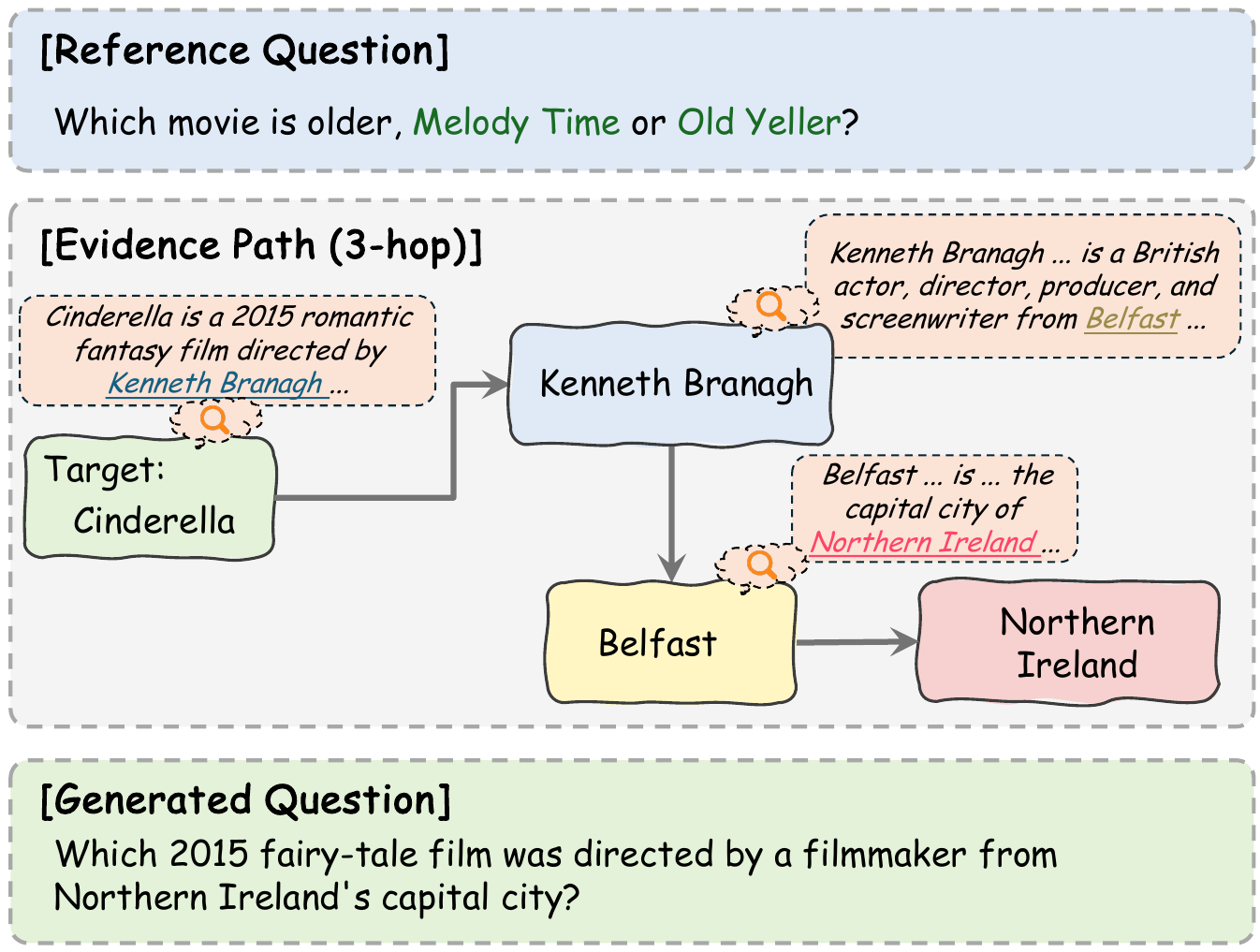}
    \caption{A case study of evidence-path guided generation.}
    \label{fig:qa_generation_case}
\end{figure}
\paragraph{Task Generation Case Study.}
Figure~\ref{fig:qa_generation_case} illustrates how \textsc{ZEST-Search} constructs a multi-hop search task from an unlabeled reference question. Instead of transforming the reference question, the proposer identifies the expected answer type and generates a target entity, Cinderella. Starting from this target, the framework constructs a source-supported evidence path connecting it to its director Kenneth Branagh, then to Belfast and Northern Ireland. Conditioned on this path, the proposer generates a multi-hop question requiring the solver to recover the target through intermediate evidence. This example demonstrates that evidence-path guidance encourages compositional question generation while maintaining verifiable answer grounding, enabling \textsc{ZEST-Search} to create challenging search tasks without labeled QA pairs.


%% file: section/related_work.tex

\section{Related Work}

\paragraph{Search-Augmented Language Models.}
Retrieval-augmented language models use external knowledge to improve generation on knowledge-intensive tasks
\citep{guu2020realm,karpukhin2020dpr,lewis2020rag}.
Later systems integrate retrieval directly into reasoning, enabling models to issue queries, inspect evidence, and refine their answers
\citep{nakano2021webgpt,yao2023react,asai2024selfrag,trivedi2023ircot}.
Search-agent reinforcement learning further optimizes when and how to search, initially through final-answer rewards and later through search-cost modeling, retrieval simulation, process supervision, and step-wise feedback
\citep{jin2025searchr1,song2025r1searcher,chen2025research,shi2025autorefine,sun2025zerosearch,zheng2025stepsearch,jiang2025s3,ji2025evorag}.
Unlike these methods, which mainly optimize solver behavior on given questions, \textsc{ZEST-Search} also brings evidence-grounded task construction into the training loop and explicitly optimizes trajectory-level search efficiency.

\paragraph{Proposer--Solver Training for Search Agents.}
More broadly, LLM self-improvement methods generate training data online, with some casting task generation and solving as interacting roles
\citep{wang2023selfinstruct,zhao2025absolutezero,chen2025sqlm,zhang2025pasodoble,creus2026populora}.
For search-agent training, EvolveSearch, Search Self-Play, and Dr.~Zero adapt generated tasks according to solver feedback, allowing task difficulty to evolve during training
\citep{zhang2025evolvesearch,lu2025search,yue2026drzero}.
Recent work further incorporates structured evidence or finer-grained process signals. For example, knowledge-graph paths can guide question generation and provide waypoint rewards, while source-level contribution scores can assign credit to retrieved evidence
\citep{wu2026knowledgegraph,arai2026eveagent}.
In contrast, \textsc{ZEST-Search} generates retrieval-supported target entities and source-supported textual evidence paths online, rather than restricting task construction to fixed answer pools or precomputed graph paths. It also decouples the proposer and solver policies and coordinates their co-evolution through objectives for intermediate task difficulty and search efficiency.

%% file: section/conclusion.tex

\section{Conclusion}
We introduced \textsc{ZEST-Search}, a zero-label framework for
tool-efficient search-agent training through heterogeneous
proposer--solver co-evolution. Rather than relying on predefined answer
pools or labeled QA pairs, \textsc{ZEST-Search} constructs validated
tasks online through answer-first generation and source-supported
multi-hop evidence paths. Lightweight, non-generative components perform
target validation, path construction, and task filtering without
additional LLM inference. Decoupling the two roles allows lightweight,
instruction-tuned, and cross-family proposers to generate complementary
tasks around the solver's evolving capability boundary. Role-specific
objectives coordinate their co-evolution by steering the proposer toward
intermediate-difficulty tasks and rewarding correct solver trajectories
with lower relative search costs.
Experiments on multiple benchmarks show that \textsc{ZEST-Search} improves answer accuracy and tool efficiency without labeled searching data. The framework remains effective with smaller and cross-family proposers, while analyses validate the reward design and task generation. Overall, \textsc{ZEST-Search} provides a scalable approach to expanding the training frontier and optimizing search efficiency.

%% file: section/appendix.tex

\appendix

\section{Appendix A: Implementation details}
\label{sec:appendix_configuration}

This section summarizes the training configurations, including general model-training parameters and method-specific parameters.
We follow Search-R1~\cite{jin2025searchr1} for the default search-related configurations, including the retriever, search protocol, and other implementation details.

\begin{table}[ht]
    \centering

    \footnotesize
    \setlength{\tabcolsep}{4pt}
    \renewcommand{\arraystretch}{1.02}
    \begin{tabular*}{\columnwidth}{
        @{\extracolsep{\fill}}lr@{}
    }
        \toprule
        \textbf{Parameter} & \textbf{Value} \\
        \midrule
        \multicolumn{2}{l}{\textit{(A) General model-training parameters}} \\
        Actor learning rate & \(3\times10^{-7}\) \\
        Proposer KL & 0.005 \\
        Solver KL & 0.001 \\
        Train batch & 128 \\
        Rollout nums & 5 \\
        Start-context tokens & 2,048 \\
        Observation tokens & 1,024 \\
        Response tokens & 1,024 \\
        Temperature & 1.0 \\
        Top-\(p\) & 0.95 \\
        Top-\(k\) & \(-1\) \\
        Training steps & 600 \\
        Retriever & E5-base-v2 \\
        Retrieval corpus & Wikipedia (Dec.~20, 2018) \\
        GPU & Nvidia A100$\times$16 \\
        \midrule
        \multicolumn{2}{l}{\textit{(B) Method-specific parameters}} \\
        Initial QA-buffer load cap & 200,000 \\
        Generated-QA extra capacity & 20,000 \\
        Solver retrieval top-\(k\) & 3 \\
        Maximum target hops & 3 \\
        Max search-enabled turns & 3 \\
        Evidence retrieval top-\(k\) & 8 \\
        Path sample top-\(N\) & 8 \\
        Path sample width & 2 \\
        Question paths/answer & 1 \\
        Random top-path pool & 4 \\
        Tool penalty coefficient $\lambda$ & 0.1 \\
        Tool-reward warm-up steps & 20 \\
        Invalid-info penalty $\beta_{\mathrm{invalid}}$ & \(-1.0\) \\
        \bottomrule
    \end{tabular*}
    \caption{Training, generation, and retrieval configurations.}
    \label{tab:appendix_training_config}
\end{table}

\newpage
\section{Appendix B: Additional Analyses}
\label{sec:appendix_analyses}

This section presents additional experimental results on frozen proposers, multi-hop question construction, and comparisons with existing self-play approaches.

\begin{table*}[t]
\centering

{\small
\setlength{\tabcolsep}{1.5pt}
\renewcommand{\arraystretch}{1}

\begin{tabular}{l@{\hspace{15pt}}l@{\hspace{10pt}}*{9}{l}}
\toprule

\multicolumn{2}{l}{}
& \multicolumn{3}{c}{\textbf{GeneralQA}}
& \multicolumn{4}{c}{\textbf{Multi-HopQA}}
&  \\
\cmidrule(lr){3-5}
\cmidrule(lr){6-9}

\multicolumn{1}{c}{\textbf{\textcolor{propcolor}{Proposer} / }\textbf{\textcolor{solvercolor}{Solver}}}
& \multicolumn{1}{c}{\textbf{Method}}
& \multicolumn{1}{c}{NQ}
& \multicolumn{1}{c}{TriviaQA}
& \multicolumn{1}{c}{PopQA}
& \multicolumn{1}{c}{HotpotQA}
& \multicolumn{1}{c}{2Wiki}
& \multicolumn{1}{c}{MuSiQue}
& \multicolumn{1}{c}{Bamboogle}
& \multicolumn{1}{c}{\textbf{Avg$(\uparrow)$}}
& \multicolumn{1}{c}{\textbf{Queries$(\downarrow)$}} \\

\midrule
\multicolumn{11}{c}{\textbf{Training Solvers with Different Proposer Configurations}} \\
\midrule

\textcolor{propcolor}{Qwen2.5-7B-Instruct /}
& Solver Base
& \base{25.8}
& \base{43.3}
& \base{28.3}
& \base{15.8}
& \base{9.2}
& \base{6.4}
& \base{24.1}
& \base{21.8}
& \base{1.59} \\

\textcolor{solvercolor}{Qwen2.5-7B-Base}
& \textbf{Ours (3-Hops)}
& \score{40.2}{}
& \score{64.5}{}
& \score{43.8}{}
& \score{36.3}{}
& \score{35.4}{}
& \score{12.0}{}
& \score{35.2}{}
& \avgscore{38.2}{16.4}
& \callscore{1.16}{-0.43} \\

\noalign{\vskip 1pt}
\cdashline{1-11}
\noalign{\vskip 1pt}

\textcolor{propcolor}{Qwen2.5-7B-Instruct/}
& Solver Base
& \base{25.8}
& \base{43.3}
& \base{28.3}
& \base{15.8}
& \base{9.2}
& \base{6.4}
& \base{24.1}
& \base{21.8}
& \base{1.59} \\

\textcolor{solvercolor}{Qwen2.5-7B-Base}
& \textbf{Frozen}
& \score{40.5}{}
& \score{61.5}{}
& \score{42.6}{}
& \score{32.8}{}
& \score{29.4}{}
& \score{9.1}{}
& \score{\textcolor{black!65}{\underline{20.0}}}{}
& \avgscore{33.7}{11.9}
& \callscore{1.48}{-0.11} \\

\noalign{\vskip 1pt}
\cdashline{1-11}
\noalign{\vskip 1pt}

\textcolor{propcolor}{Qwen2.5-7B-Instruct}
& Solver Base
& \base{25.8}
& \base{43.3}
& \base{28.3}
& \base{15.8}
& \base{9.2}
& \base{6.4}
& \base{24.1}
& \base{21.8}
& \base{1.59} \\

\textcolor{solvercolor}{Qwen2.5-7B-Base}
& \textbf{4 Target Hops}
& \score{41.0}{}
& \score{64.3}{}
& \score{43.3}{}
& \score{37.4}{}
& \score{35.8}{}
& \score{14.1}{}
& \score{39.2}{}
& \avgscore{39.3}{17.5}
& \callscore{1.28}{-0.31} \\

\midrule
Qwen2.5-7B-Instruct
& \textbf{Dr.Zero}
& \base{\textcolor{black!65}{\underline{40.6}}}
& \base{\textcolor{black!65}{\underline{60.8}}}
& \base{\textcolor{black!65}{\underline{41.6}}}
& \base{\textcolor{black!65}{\underline{36.2}}}
& \base{\textcolor{black!65}{\underline{34.7}}}
& \base{\textcolor{black!65}{\underline{10.4}}}
& \base{\textcolor{black!65}{\underline{36.0}}}
& \base{\textcolor{black!65}{\underline{37.2}}}
& \base{-} \\

\bottomrule
\end{tabular}
}

\caption{Additional experiment results. Parenthesized numbers are training steps. Results above the base model are in bold. The results below the base model or below our method are in gray and underlined.}
\label{tab:additional_ablation_results}
\end{table*}

\subsection{B.1 Additional Ablation Study}
\label{sec:appendix_additional_ablation}

\paragraph{Frozen-Proposer Control.}
\begin{figure}
    \centering
    \includegraphics[width=\linewidth]{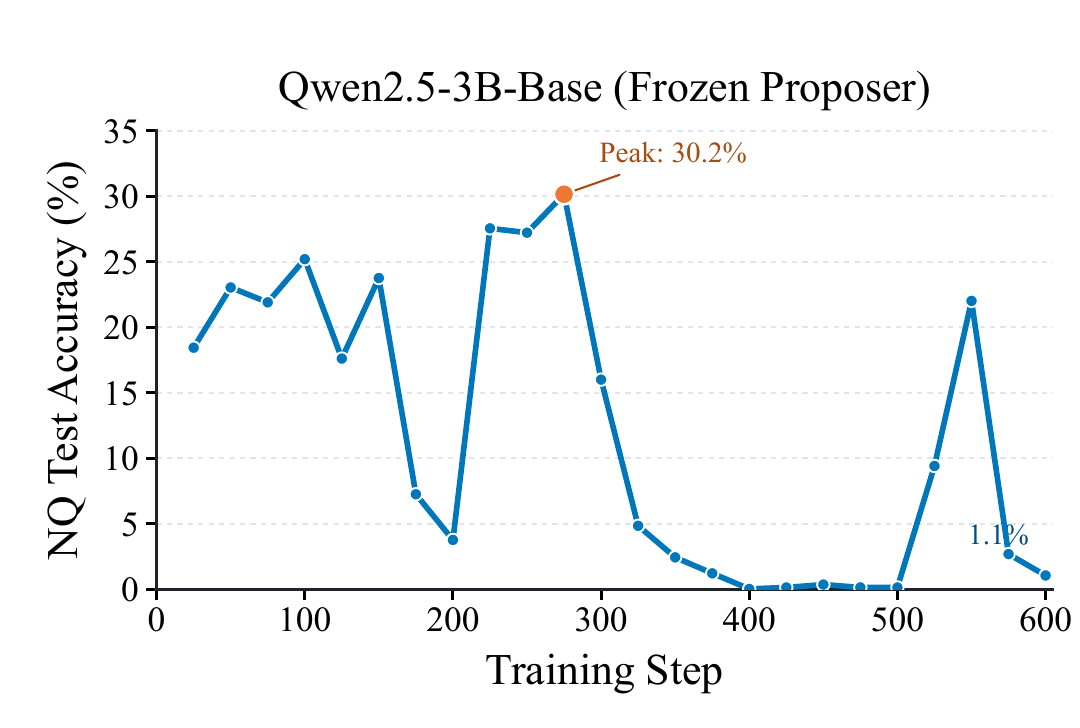}
    \caption{Training-time test accuracy of \textsc{ZEST-Search} with a frozen Qwen2.5-3B-Base proposer and a Qwen2.5-3B-Base solver}
    \label{fig:3b_frozen}
\end{figure}

As shown in Table~\ref{tab:additional_ablation_results}, Qwen2.5-7B-Instruct, as an instruction-tuned model, can generate relatively high-quality questions through evidence-path guided generation, leading to improvements on general QA tasks. However, without co-evolution with the solver, the fixed proposer struggles to continuously generate high-quality multi-hop tasks aligned with the solver's evolving capability boundary, resulting in limited gains and even performance degradation on some multi-hop benchmarks.
For base models with weaker instruction-following ability, such as a frozen Qwen2.5-3B-Base proposer shown in Figure~\ref{fig:3b_frozen}, the quality of generated tasks is more constrained by the proposer itself. Since task validation relies on lightweight NLP components rather than additional LLM-based filtering, low-quality tasks may accumulate when the proposer cannot adapt to the solver. This can weaken the training signal and lead to degraded solver performance.

Together, these results demonstrate the effectiveness and necessity of proposer--solver co-evolution for generating effective training tasks, adapting to the solver's evolving capability, and improving search-agent training.

\begin{table}[t]
    \centering
    \small
    \setlength{\tabcolsep}{4pt}
    \renewcommand{\arraystretch}{1.05}
    \begin{tabular}{lrr}
        \toprule
        \textbf{Stage}
        & \textbf{Accepted}
        & \textbf{Rate (\%)} \\
        \midrule
        Target format and entity filtering
        & 486/500 & 97.2 \\
        Retrieval-support verification
        & 411/486 & 84.6 \\
        \textbf{Overall target validation}
        & \textbf{411/500} & \textbf{82.2} \\
        Evidence-path construction
        & 411/411 & 100.0\textsuperscript{*} \\
        Static QA verification
        & 302/411 & 73.5 \\
        \textbf{Final task acceptance}
        & \textbf{302/500} & \textbf{60.4} \\
        \bottomrule
    \end{tabular}
    \caption{Acceptance rates across the EPG filtering stages.
    \textsuperscript{*}Path construction is evaluated only on targets
    that pass retrieval-support verification.}
    \label{tab:epg_acceptance}
\end{table}

\begin{table}[ht]
    \centering
    \small
    \setlength{\tabcolsep}{3pt}
    \renewcommand{\arraystretch}{1.05}
    \begin{tabular}{lrr}
        \toprule
        \textbf{Stage}
        & \textbf{Total (s)}
        & \textbf{Per Task (s)} \\
        \midrule
        Target generation and validation
        & 142.53 & 0.472 \\
        Evidence-path construction
        & 59.24 & 0.196 \\
        Question generation and static verification
        & 111.15 & 0.368 \\
        \midrule
        \textbf{Full pipeline}
        & \textbf{312.92} & \textbf{1.036} \\
        \bottomrule
    \end{tabular}
    \caption{Runtime breakdown of EPG over 500 reference questions.
    Per-task runtime is amortized over 302 accepted tasks with a
    generation batch size of 64.}
    \label{tab:epg_runtime}
\end{table}

\paragraph{Ablation on Multi-Hop Task Complexity.}

We further investigate whether increasing the complexity of generated tasks improves the solver's multi-hop search capability. Specifically, we increase the maximum target path length to 4, allowing EPG to construct evidence paths containing up to four entities. To provide the solver with sufficient interaction budget for resolving these longer paths, we also increase the maximum number of search turns to 4, while keeping all other training and evaluation configurations unchanged. This controlled setting isolates the effect of longer evidence paths and more demanding multi-hop supervision.

As shown in Table~\ref{tab:additional_ablation_results}, generating higher-hop tasks provides limited improvements on general QA benchmarks but leads to more substantial gains on multi-hop QA tasks. This suggests that the proposed evidence-path guided generation can effectively construct training tasks with stronger compositional reasoning requirements, enabling the solver to acquire genuine multi-hop search capabilities rather than merely improving single-step retrieval or answer matching.

\subsection{B.2 EPG Runtime}
\label{sec:appendix_epg_runtime}

\paragraph{Efficiency and Acceptance Analysis of EPG.}

We analyze the filtering behavior and runtime efficiency of the
evidence-path guided generation (EPG) pipeline using 500 reference
questions generated by a Qwen2.5-7B-Instruct proposer. As shown in
Table~\ref{tab:epg_acceptance}, 82.2\% of the generated target entities
pass target validation, and 302 questions remain after static QA
verification, yielding an overall task acceptance rate of 60.4\%.
Notably, evidence paths can be constructed for all retrieval-supported
targets without additional LLM-based verification or inference.

Table~\ref{tab:epg_runtime} reports the runtime breakdown of EPG. 
Evidence-path construction takes only 59.24 seconds, substantially less than target generation and question generation, which require 142.53 and 111.15 seconds, respectively. Unlike prior approaches that rely on an LLM judge or additional reasoning calls to validate generated questions~\cite{lu2025search, zhang2025evolvesearch, wu2026knowledgegraph}, our pipeline uses only lightweight retrieval and NLP components. It therefore avoids extra LLM inference, reduces validation overhead, and efficiently filters generated tasks while preserving source-supported evidence for multi-hop question generation.

\begin{figure}[ht]
    \centering
    \includegraphics[width=\linewidth]{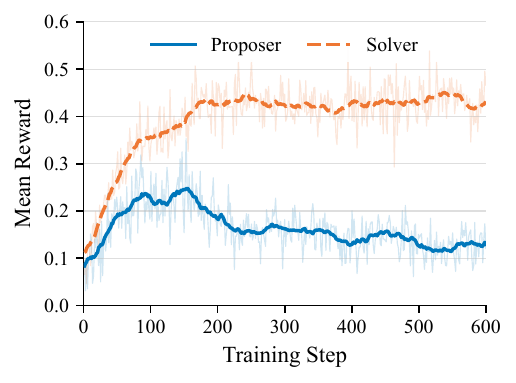}
    \caption{Training reward dynamics of the Qwen2.5-7B-Instruct proposer and Qwen2.5-7B-Base solver.}
    \label{fig:reward-dynamics}
\end{figure}

\subsection{B.3 Reward Dynamics}
\label{sec:appendix_reward_dynamics}

\paragraph{Reward Dynamics.}

Figure~\ref{fig:reward-dynamics} shows distinct but coupled reward dynamics for the proposer and solver. The solver reward rises rapidly during early training and then stabilizes at a high level, indicating that the solver progressively acquires and maintains correct, tool-efficient search behavior. In contrast, the proposer reward first increases and then gradually decreases to a stable positive range. This decline should not be interpreted as degraded proposer performance. Under the intermediate-difficulty objective, questions that are initially informative become easier as the solver improves, reducing their proposer reward and continuously shifting the preferred task distribution toward the solver's evolving capability boundary. The sustained positive proposer reward suggests that the proposer continues to generate useful frontier tasks, while the stable solver reward indicates that this adaptive task generation does not destabilize solver optimization. Short-term fluctuations arise naturally from stochastic online task generation and grouped rollouts. Since the two roles use different reward definitions, their absolute reward values are not directly comparable; the key observation is their stable and coordinated evolution over training.

\subsection{B.4 Comparison with Self-Play Methods.}
A strictly controlled comparison with existing self-play methods is
difficult because they differ substantially in task sources, validation
mechanisms, and search protocols. 
Many methods rely on predefined supports, such as answer pools~\cite{lu2025search} or knowledge graphs~\cite{wu2026knowledgegraph}, and additionally use LLM-based judging or reasoning to verify the validity and correctness of generated questions and answers. Such verification is sensitive to the choice of judge model, and different LLMs may produce different filtering decisions and results. Moreover, these methods did not release trained checkpoints, making it difficult to reevaluate them under a unified search protocol and evaluation standard.
We therefore compare with
the reported results of Dr.~Zero~\cite{yue2026drzero}, which is closest to our zero-label
setting, while treating this comparison as indicative rather than fully
controlled. Dr.~Zero uses Qwen2.5-7B-Instruct for both roles and allows
up to five search turns, whereas \textsc{ZEST-Search} trains a
Qwen2.5-7B-Base solver with at most three search turns. Despite using a
base rather than instruction-tuned solver and a smaller search budget,
our default 3-hop setting improves the average score from 37.2 to 38.2.
Increasing the target path length to four hops further raises the average
to 39.3 and outperforms Dr.~Zero on all seven benchmarks. 
As reported in the original paper, Dr.~Zero improves over Search-R1 on
several individual datasets but achieves a slightly lower average
performance. In contrast, our main results demonstrate substantially
larger improvements over Search-R1 and ZeroSearch~\cite{sun2025zerosearch} without any labeled search data.

\section{Appendix C: Lightweight Evidence-Path Construction and
Static Question Validation}
\label{sec:appendix_epg_implementation}

\subsection{C.1 Overview}

This section describes the implementation of the evidence-path guided
question construction pipeline used in our training framework. 
The goal of this module is to automatically construct corpus-supported
candidate questions from unlabeled references without invoking additional
LLM-based judges or human annotation.
Given a reference question, the pipeline first generates
a searchable target entity, verifies its existence through dense retrieval,
constructs an evidence-supported entity path using lightweight NLP
components,
and finally generates and validates a new question.

Unlike approaches that employ an additional language model or NLI-based
verifier to assess generated questions, our validation procedure relies on
retrieval grounding, entity-level consistency, and deterministic rules. This
design significantly reduces verification overhead while maintaining reliable
control over the generated training distribution.

The complete pipeline consists of four stages:
(1) target entity generation and retrieval-based verification,
(2) lightweight evidence-path construction,
(3) evidence-path guided question generation,
and (4) static question validation before solver rollout.

\subsection{C.2 Retrieval-based Target Verification}
\label{sec:appendix_target_verification}

\paragraph{Dense retrieval backbone.}
All retrieval operations in the evidence construction module use the same
dense retrieval service. Specifically, we adopt
\texttt{intfloat/e5-base-v2} with a pre-built FAISS index over the
Wikipedia-2018 corpus. Each query is encoded with the required
\texttt{query:} prefix, truncated to 256 tokens, mean-pooled using the
attention mask, and $\ell_2$ normalized before nearest-neighbor search.

For target verification and evidence expansion, the retriever returns the
top-$8$ passages. This setting is independent from the solver search
environment, where each search action retrieves top-$3$ passages. The
retrieval results are only used as corpus grounding signals and do not
directly determine answer correctness.

\paragraph{Target entity generation.}
The first proposer call does not answer the reference question. Instead, it
is instructed to infer the semantic type of the expected answer and generate
a different searchable entity with a similar semantic category. For example,
if the reference question expects a person, organization, location, or work,
the proposer generates another entity belonging to the same category. For
questions whose answers are numerical or temporal values, the proposer
identifies the corresponding entity or topic that determines such values and
uses it as the target anchor.

The generated target $a^\star$ must satisfy two requirements:

\begin{enumerate}
    \item it should represent a concrete searchable entity;
    \item it should be supported by the retrieval corpus.
\end{enumerate}

Before retrieval verification, candidates undergo deterministic surface
filtering to remove malformed outputs, including empty strings, placeholders,
non-entity values, URLs, question-like generations, and residual generation
tags.

\paragraph{Retrieval support verification.}
A candidate is considered retrieval-supported if its normalized form
satisfies at least one of the following conditions: (i) it exactly matches a
retrieved title; (ii) it and a retrieved title are mutually contained, with
both normalized strings containing at least four characters; (iii) its
token-set Jaccard similarity with a retrieved title is at least $0.8$; or
(iv) it contains at most six tokens and appears within the first 1,200
characters of a retrieved passage.

This stage only verifies that the target entity has corpus support. It does
not perform entity linking, factual verification, or relation verification.

\subsection{C.3 Evidence Path Construction}
\label{sec:appendix_path_construction}

After obtaining a retrieval-supported target entity $a^\star$, we construct
a lightweight evidence graph from retrieved passages. The objective is not
to build a complete knowledge graph, but to extract local evidence chains
that provide controllable generation context for the proposer.

The evidence graph is represented as:

\[
G=(V,E),
\]

where each node $v\in V$ corresponds to a normalized entity mention and each
edge $e\in E$ denotes a sentence-level entity association supported by
co-mention and lexical cues. Each edge stores its source passage, verbatim
supporting sentence, entity endpoints, and lexical support indicators.

\subsubsection{Entity Extraction and Normalization}

Entity extraction combines lightweight NLP tools and deterministic rules.
When spaCy NER is available, we retain mentions labeled
\texttt{PERSON}, \texttt{ORG}, \texttt{GPE}, \texttt{LOC},
\texttt{FAC}, \texttt{WORK\_OF\_ART}, \texttt{EVENT}, and
\texttt{PRODUCT}.
Numeric and temporal mentions are removed because they are less suitable
for constructing stable entity-centered paths.

When a document title appears in a sentence, it is treated as a
high-confidence entity mention. A capitalization-based extractor is used only
as a fallback when no valid NER entity is detected.

Unlike external entity-linking systems such as Wikidata-based linking, our
method performs local within-corpus normalization. Entity mentions are
normalized by Unicode normalization, case folding, punctuation removal, and
surface matching. Each entity receives a deterministic identifier derived
from its canonical representation.

This lightweight design avoids dependency on external knowledge bases while
providing sufficient consistency for evidence-path construction.

\subsubsection{Evidence Edge Construction}

Starting from the target entity $a^\star$, retrieved passages are segmented
into sentences. If a sentence contains both $a^\star$ and another valid
entity $v$, we construct a directed edge:

\[
v \rightarrow a^\star .
\]

The edge direction follows the reverse traversal order of question
construction, where the final node is always the target answer.
Each edge preserves the original evidence sentence and its document
metadata. During expansion, the current frontier entity is used as a new
retrieval query. Newly discovered entities form additional edges:

\[
v_i \rightarrow v_{i+1}.
\]

To avoid shortcut paths, documents containing the final answer entity are
excluded after the initial seed retrieval stage.

Edge quality is determined through deterministic lexical signals rather than
a learned relation extraction model. Explicit relational patterns produce
strong edges, while weaker co-occurrence patterns receive lower confidence.

Candidate edges are ranked lexicographically by relation strength, title
support, entity confidence, ascending retrieval rank, and descending
retrieval score.
This deterministic priority order favors better-supported candidate edges
without introducing an additional neural relation extractor.

\subsubsection{Path Sampling and Certification}

The evidence-path builder expands the graph iteratively from the target
entity. Instead of exhaustively enumerating all possible paths, which can
lead to combinatorial growth, we adopt a lightweight sampling-based
expansion strategy.

At each expansion step, candidate entities are ranked according to the edge
quality score. Among the top-$8$ candidates, the highest-ranked candidate is
always preserved, while one additional candidate is randomly sampled with a
fixed training-step-dependent seed. Therefore, the default branching width is
two. With a maximum hop budget of three, each target produces at most
$2^3=8$ candidate paths before certification.

The hop budget is treated as an upper bound rather than a fixed requirement.
When no reliable frontier entity can be expanded, the current shorter path is
retained. This allows the system to avoid introducing unsupported relations
while maintaining sufficient generation coverage.

Each candidate path is represented as:

\[
P=(v_0,e_1,v_1,\ldots,e_h,v_h),
\]

where $v_h=a^\star$ is the target answer entity. Paths satisfying the
following constraints are marked as certified:

\begin{table}[t]
\centering
\small
\setlength{\tabcolsep}{4pt}
\begin{tabularx}{\columnwidth}{@{}p{0.32\columnwidth}X@{}}
\toprule
\textbf{Criterion} & \textbf{Requirement} \\
\midrule
Endpoint &
The final node corresponds to the target answer $a^\star$. \\

Continuity &
Adjacent edges have consistent entity endpoints. \\

Acyclicity &
No entity identifier is repeated within a path. \\

Evidence grounding &
Each supporting sentence occurs verbatim in its source document. \\

Entity support &
Each evidence sentence contains both edge endpoints. \\

Relation validity &
Each edge satisfies the specified lexical relation constraints. \\
\bottomrule
\end{tabularx}
\caption{Evidence-path structural certification criteria.}
\label{tab:path_certification}
\end{table}

Expansion may terminate before reaching the maximum hop budget, in which
case the current shorter path is retained. If no supported edge is found,
the retrieved answer passages form an answer-only grounding context. If no
candidate passes the structural checker, the highest-ranked candidate may
still be retained as an explicitly logged uncertified fallback. Accordingly,
context availability should not be conflated with successful path
certification.

Importantly, path certification only measures structural and lexical
consistency with retrieved corpus evidence. It does not perform external fact
checking, natural language inference, or LLM-based judgment. 
Therefore, certified paths should be interpreted as
retrieval-supported generation contexts rather than formally verified
knowledge representations.

\subsection{C.4 Evidence-path Guided Question Generation}
\label{sec:appendix_question_generation}

After path construction, one evidence path is selected for each target entity
and provided to the proposer for question generation. The input context
contains:

\begin{itemize}
    \item the target answer entity $a^\star$;
    \item the ordered entity chain from the outer entity to the target;
    \item intermediate bridge entities;
    \item supporting evidence sentences;
    \item source document information and relation indicators.
\end{itemize}

Conditioned on the target entity, selected path, and supporting documents,
the proposer generates a question whose answer is expected to be exactly
$a^\star$.
The generation objective is to produce a self-contained question whose answer
is exactly $a^\star$, while avoiding direct leakage of the answer entity.
The proposer is encouraged to utilize multiple evidence steps when supported
by the context, but shorter questions are allowed when a shorter composition
provides clearer and more reliable evidence.

The generation prompt specifies two desiderata: the question should be
consistent with the supplied evidence context, and the target answer should
not appear explicitly in the question. The deterministic validator directly
enforces the second requirement through normalized alias matching. The first
is encouraged through evidence-conditioned generation but is not separately
verified by an entailment model or an LLM judge.

The complete evidence path is therefore used as a controllable generation
signal rather than merely as additional retrieval information.

\subsection{C.5 Static Question Validation}
\label{sec:appendix_question_validation}

Before entering solver rollout, generated questions undergo a deterministic
validation procedure. The purpose of this stage is to remove malformed
generations and prevent trivial answer leakage.

The validator operates only on the generated question and the serialized
materials produced during evidence construction.

The validation criteria are summarized in
Table~\ref{tab:question_validation}.

\begin{table}[t]
\centering
\small
\setlength{\tabcolsep}{3pt}
\renewcommand{\arraystretch}{1.05}
\begin{tabularx}{\columnwidth}{
    @{}
    p{0.20\columnwidth}
    p{0.33\columnwidth}
    X
    @{}
}
\toprule
\textbf{Criterion} &
\textbf{Check} &
\textbf{Purpose} \\
\midrule

Format &
Valid question block and non-empty content &
Ensure parseability. \\

Length &
Character- and word-level constraints &
Remove malformed outputs. \\

Support &
Non-empty evidence materials &
Ensure grounding is available. \\

Leakage &
Normalized target-alias matching &
Prevent trivial answer exposure. \\

Entity exposure &
Optional path-based entity checks &
Control entity exposure. \\

\bottomrule
\end{tabularx}
\caption{Static question validation criteria.}
\label{tab:question_validation}
\end{table}

Specifically, the validator performs the following checks:

\paragraph{Output format.}
The final non-empty question block is extracted from the proposer output.
Missing or empty question generations are rejected.

\paragraph{Length constraint.}
Questions shorter than 10 characters or longer than 200 characters are
removed. A minimum whitespace-separated word count is also applied to avoid
degenerate outputs.

\paragraph{Answer leakage detection.}
The validator constructs multiple normalized forms of the target answer,
including:

\begin{itemize}
    \item original surface form;
    \item punctuation-normalized form;
    \item compact alphanumeric form;
    \item case-folded representation.
\end{itemize}

A generated question is rejected if any normalized target form appears in the
question text.

\paragraph{Evidence availability.}
A question is accepted only when the serialized evidence materials are
non-empty. The validator does not independently retrieve new documents or
perform additional semantic evaluation.

Consequently, static validation ensures that a question is structurally
well formed, accompanied by non-empty evidence materials, and free from
direct target leakage under the implemented alias-matching rules. It does
not establish evidence entailment, answer uniqueness, factual correctness,
or multi-hop necessity.

\subsection{C.6 Computational Efficiency}
\label{sec:appendix_efficiency}

The evidence construction and validation stages introduce no additional LLM
inference beyond the two proposer calls used for target and question
generation. As quantified in Table~\ref{tab:epg_runtime}, evidence-path
construction requires 59.24 seconds over 500 reference questions, compared with 142.53 seconds for target generation and validation and 111.15 seconds for question generation and static verification. 
Thus, the non-generative evidence stage contributes substantially less wall-clock time than either LLM-generation stage while preserving explicit evidence provenance for each accepted candidate.

Static acceptance is deliberately separated from solver-based evaluation.
The former removes malformed candidates and direct answer leakage before
rollout, whereas the latter supplies dynamic correctness, difficulty, and
tool-use signals only after acceptance. This ordering avoids spending solver
rollouts on candidates that already violate inexpensive deterministic
constraints.